\documentclass[conference]{IEEEtran}
\IEEEoverridecommandlockouts
\usepackage{cite}
\usepackage{amsmath,amssymb,amsfonts}
\usepackage{algorithmic}
\usepackage{graphicx}
\usepackage{textcomp}
\usepackage{xcolor}
\usepackage{multirow}
\usepackage{booktabs}
\usepackage{makecell}
\usepackage{pifont}
\usepackage{amssymb}
\usepackage{cleveref}
\usepackage{color}

\usepackage{stfloats}
\def\BibTeX{{\rm B\kern-.05em{\sc i\kern-.025em b}\kern-.08em
    T\kern-.1667em\lower.7ex\hbox{E}\kern-.125emX}}
\begin{document}

\title{HSACNet: Hierarchical Scale-Aware Consistency Regularized Semi-Supervised Change Detection}

% \author{Anonymous ICME submission}
% \author{Paper ID \#1473}
% \author{
% Qi'ao Xu\textsuperscript{†}, Pengfei Wang\textsuperscript{†}, Yanjun Li, Tianwen Qian, Xiaoling Wang\textsuperscript{\textasteriskcentered}
% }

\author{
    \IEEEauthorblockN{
    Qi'ao Xu\textsuperscript{†}, Pengfei Wang\textsuperscript{†}, Yanjun Li, Tianwen Qian\textsuperscript{\textasteriskcentered}, Xiaoling Wang\textsuperscript{\textasteriskcentered}
    }
    \IEEEauthorblockA{
        % \IEEEauthorrefmark{1}
        School of Computer Science and Technology, East China Normal University, Shanghai, 200062, China \\
        Email: 
        \{qaxu,pfwang,51265901098\}@stu.ecnu.edu.cn, \{twqian,xlwang\}@cs.ecnu.edu.cn
    }
    \thanks{$^{\ast}$Corresponding author: Tianwen Qian, Xiaoling Wang.}
    \thanks{\textsuperscript{†}These authors contributed equally.}
    \thanks{This work was supported by NSFC grant (No. 62136002 and 62477014), Ministry of Education Research Joint Fund Project (8091B042239), and Shanghai Trusted Industry Internet Software Collaborative Innovation Center.}
}

\maketitle

\begin{abstract}
Semi-supervised change detection (SSCD) aims to detect changes between bi-temporal remote sensing images by utilizing limited labeled data and abundant unlabeled data. Existing methods struggle in complex scenarios, exhibiting poor performance when confronted with noisy data. They typically neglect intra-layer multi-scale features while emphasizing inter-layer fusion, harming the integrity of change objects with different scales. In this paper, we propose HSACNet, a Hierarchical Scale-Aware Consistency regularized Network for SSCD. Specifically, we integrate Segment Anything Model 2 (SAM2), using its Hiera backbone as the encoder to extract inter-layer multi-scale features and applying adapters for parameter-efficient fine-tuning. Moreover, we design a Scale-Aware Differential Attention Module (SADAM) that can precisely capture intra-layer multi-scale change features and suppress noise. Additionally, a dual-augmentation consistency regularization strategy is adopted to effectively utilize the unlabeled data. Extensive experiments across four CD benchmarks demonstrate that our HSACNet achieves state-of-the-art performance, with reduced parameters and computational cost.
\end{abstract}


\section{Introduction}
\label{sec:intro}

Change detection (CD), a foundational task in remote sensing, focuses on identifying pixel-level changes between images captured at the same location but different times. It plays a pivotal role in various fields, such as disaster monitoring, urban development, and resource management~\cite{bai2023deep,Holail2024nsr}.
Over the past decade, fully supervised CD has become mainstream with excellent performance. But it heavily depends on vast pixel-level annotations, which are laborious and time-consuming to acquire, thus impeding its practical application~\cite{cheng2023change}.
Semi-supervised change detection~(SSCD) has drawn growing attention for enhancing model generalization by utilizing limited labeled data and abundant unlabeled data.

% motivation
\begin{figure}[!ht]
\graphicspath{{Fig/}}
\centering
{\includegraphics[width=1.0\linewidth]{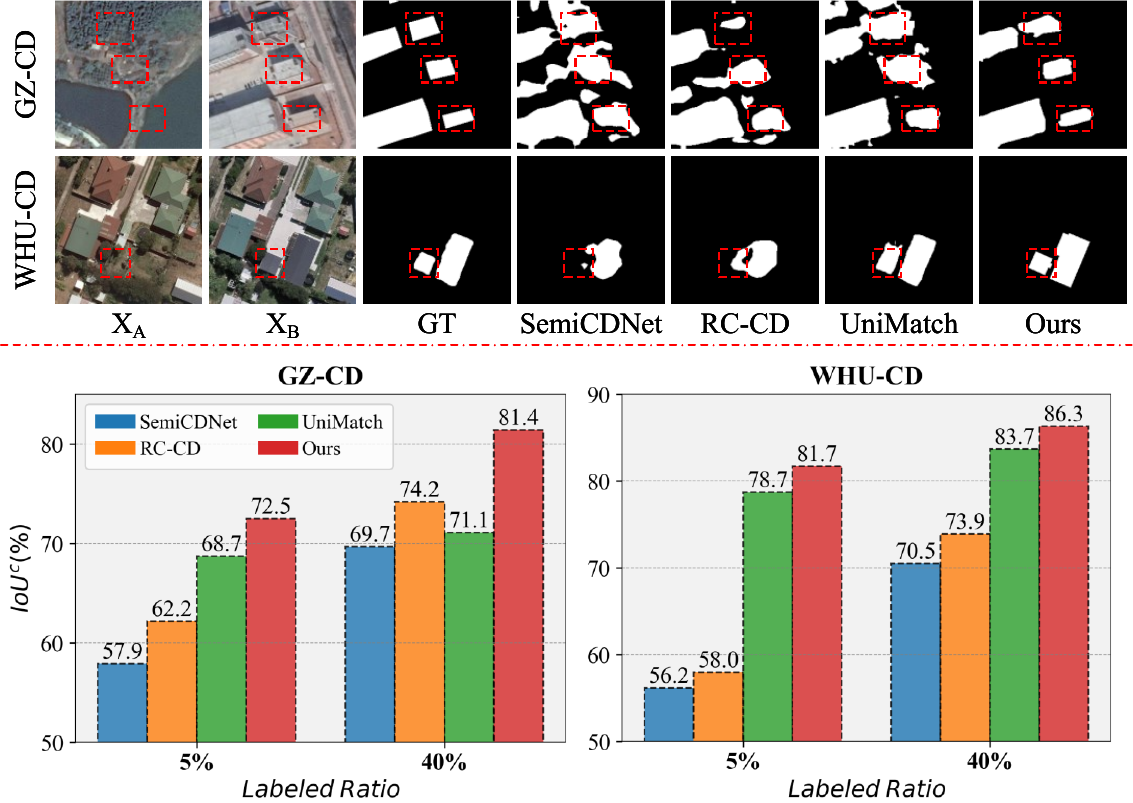}}
%	\vspace{-4px}
\caption{Motivation analysis for SSCD on GZ-CD and WHU-CD datasets. \textbf{Upper:} Visual examples at \(5\%\) labeled ratio, exposing existing methods' insensitivity to variations in scenes and scales. The objects in \textcolor{red}{red dotted box} are notable changes. \textbf{Lower:} Quantitative comparison of different methods at both \(5\%\) and \(40\%\) labeled ratios, spotlighting the need for our research. }
\label{fig:motivation}
\vspace{-8px}
\end{figure}
%

% change detection network
\begin{figure*}[!ht]
\graphicspath{{Fig/}}
\centering{\includegraphics[width=0.88\linewidth]{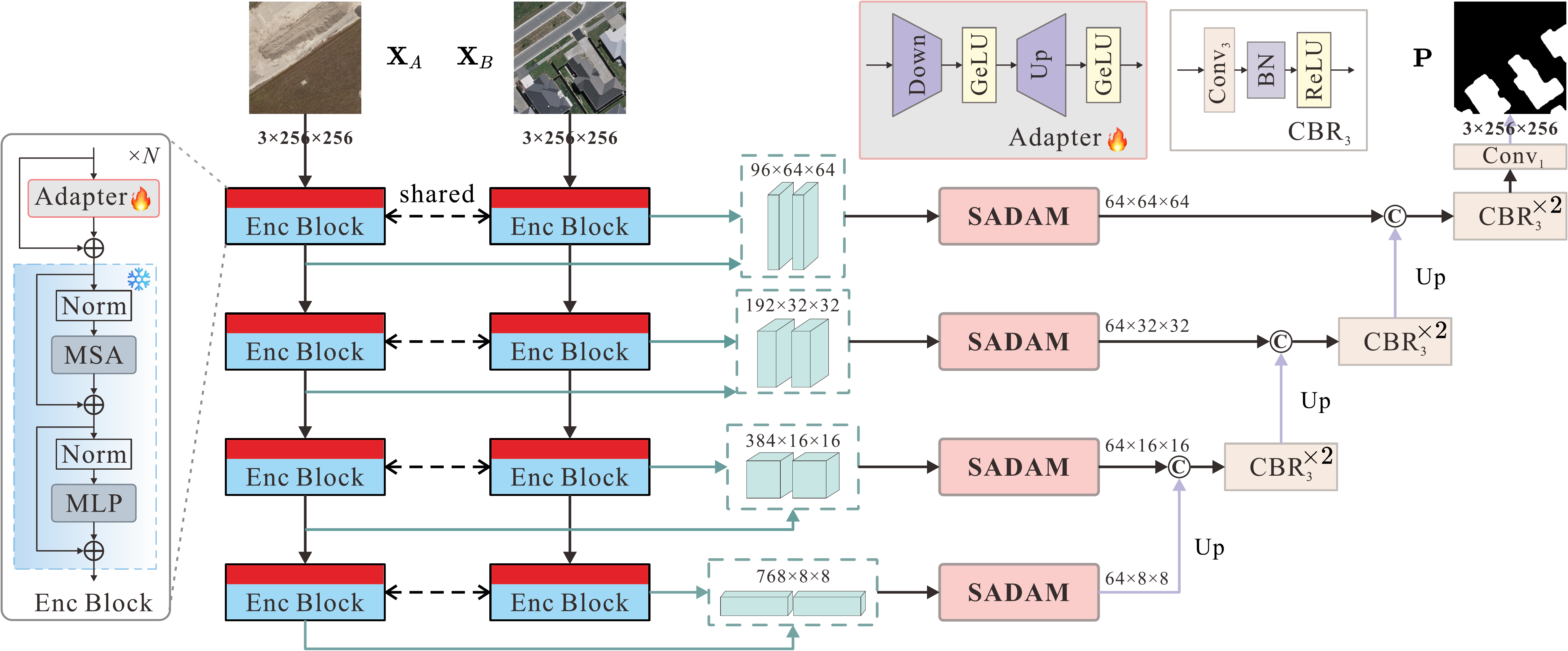}}
\vspace{-2px}
\caption{The architecture of our change detection network, which consists of three components: 1) the SAM2 encoder with adapters for multi-scale feature extraction, 2) the SADAM for refining change features via multi-branch group convolutions and spatial-channel attention, and 3) the decoder for fusing cross-scale features to generate change maps.}
\label{fig:framework}
\vspace{-4px}
\end{figure*}

Despite the notable progress, current SSCD methods still face significant challenges when applied to complex scenarios. The most prominent challenge stems from the scarcity of labeled data, which not only limits the model's generalization ability across diverse scenarios but also undermines its robustness in handling noise.
As shown in Fig.~\ref{fig:motivation}, this limitation manifests in two key aspects:
1) Remote sensing images involve variations in weather, seasons, and lighting, posing substantial challenges to the model's adaptability across diverse scenarios; 
2) Object changes occur at different scales and are accompanied by noise, reducing the model's effectiveness in handling fine-grained variations.

We argue that the underlying reasons for the poor performance lies in the low-quality change features, arising from two main factors: 
1) Existing networks possess an inadequate feature extraction ability. With few labeled samples, encoders struggle to exploit the discriminative capabilities of features, thus failing to extract precise and relevant object features. 
2) Simplistic feature operations like basic difference or naive concatenation are insufficient. They neglect complex spatial and spectral correlations, failing to capture change objects across different scales. 
As a result, the generated change features are of low quality, impeding the detection performance.

To address the above mentioned issues, we present HSACNet, a Hierarchical Scale-Aware Consistency regularized Network for SSCD.
It incorporates the Hiera backbone from Segment Anything Model 2 (SAM2)~\cite{ravi2024sam} as an encoder for inter-layer multi-scale features extraction, and uses adapters for parameter-efficient fine-tuning to optimize learning.
The Scale-Aware Differential Attention Module (SADAM) is designed, through multi-branch group convolutions and spatial-channel attention, to capture intra-layer multi-scale change features and suppress noise. 
A dual-augmentation consistency regularization is employed to enhance the utilization of unlabeled data.
We conduct experiments on four CD benchmarks, and our HSACNet achieves state-of-the-art performance with fewer parameters and lower computational cost.

Our contributions are summarized as follows:
\begin{itemize}
  \item We present an advanced SSCD method HSACNet. It applies SAM2's Hiera backbone to extract inter-layer multi-scale features and inserts adapters for parameter-efficient fine-tuning, enhancing the detection performance. 
  \item We develop a Scale-Aware Differential Attention Module (SADAM), which precisely captures intra-layer multi-scale change features and reduces noise.
  \item Extensive experiments on four CD benchmarks demonstrate the superiority and effectiveness of our HSACNet.

\end{itemize}

% % ----------------------------------------------------------------------------------------------------------------

\section{Related Work}

\textbf{Semi-supervised Change Detection.} 
Existing SSCD methods can be broadly categorized into three types. Adversarial learning-based methods use alternative optimization strategies via adversarial mechanisms for better feature learning~\cite{peng2020semicdnet}. Pseudo-label-based methods focus on selecting high-quality initial pseudo-labels, predicting unlabeled data, filtering, and iteratively retraining the network~\cite{wang2022reliable}. Consistency regularization-based methods enforce perturbed images or features to yield the same output as the original~\cite{yang2023revisiting}.

\textbf{Segment Anything Model.} 
Segment Anything Model (SAM)~\cite{kirillov2023segment} emerges as the pioneering vision foundation model designed for image segmentation. It demonstrates remarkable potential in computer vision and has widely applied across various applications. SAM2, developed from SAM1 with expanded training data and enhanced architecture, exhibiting superior performance and broader applicability.

% %－－－－－－－－－－－－－－－－－－－－－－－－－－－－－－－－－－－－－－－－－－－－－－－－－－－－－－－－－－－－－－

\section{Methodology}

\subsection{Problem Formulation}

Semi-supervised change detection~(SSCD) endeavors to generate accurate change maps by leveraging limited labeled data and abundant unlabeled data. 
The training set consists of two subsets: a labeled set and an unlabeled set. 
The labeled set is formulated as
$\mathcal{D}_l= \{{(\mathbf{X}^l_{A,i}, \mathbf{X}^l_{\mathrm{B},i}), \mathbf{Y}^l_i}\}^M_{i=1}$.
Here, $(\mathbf{X}^l_{A,i},\mathbf{X}^l_{B,i})$ represents the $i$-th labeled image pair, with $\mathbf{X}^l_{A,i}$ being the pre-change image, $\mathbf{X}^l_{B,i}$ the post-change image, and $\mathbf{Y}^l_i$ the corresponding ground truth. 
The unlabeled set is defined as $\mathcal{D}_u=\{{(\mathbf{X}^u_{A,i}, \mathbf{X}^u_{B,i})}\}^N_{i=1}$, where $(\mathbf{X}^u_{A,i},\mathbf{X}^u_{B,i})$ is the $i$-th unlabeled image pair. 
$M$ and $N$ denote the number of labeled and unlabeled image pairs, respectively. 
In most scenarios, it holds that $N>>M$.
The following sections will detail the proposed HSACNet.

\subsection{Change Detection Network}

As shown in Fig.~\ref{fig:framework}, our CD network $\phi$ comprises three key components: the SAM2 image encoder with adapters, the scale-aware differential attention module, and the decoder.

\textbf{SAM2 Image Encoder with Adapters.}
Given the powerful feature extraction capability of SAM2, we select its Hiera-T~\cite{ryali2023hiera} backbone with 28M parameters as the basis for our siamese encoder. It has 4 layers with channels $=[96,192,384,768]$ and blocks $=[1,2,7,2]$. This hierarchical structure excels at extracting inter-layer multi-scale features from input image pairs. The process is expressed as:
\begin{equation}
	\mathbf{C}^A_{i}, \mathbf{C}^B_{i} = \phi_{Enc}(\mathbf{X}_{A}, \mathbf{X}_{B}), i=1,2,3,4 
\end{equation}
where $\mathbf{X}_{A}$ and $\mathbf{X}_{B}$ are pre- and post-change images, $\mathbf{C}^A_{i}$ and $\mathbf{C}^B_{i}$ are the corresponding multi-scale features.
%

%  SADAM
\begin{figure}[!ht]
\graphicspath{{Fig/}}
\centering
{\includegraphics[width=0.95\linewidth]{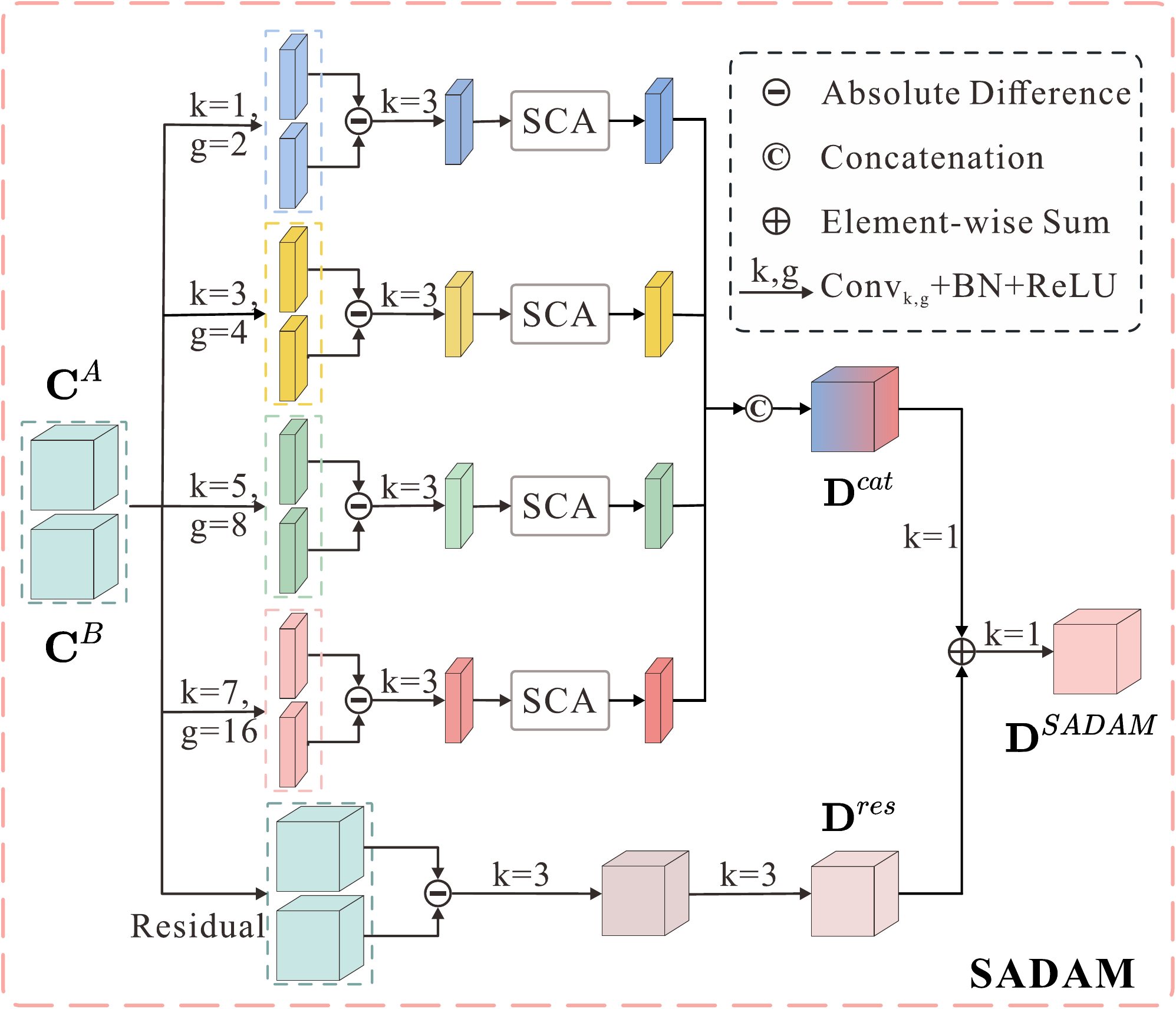}}
% \vspace{-2px}
\caption{Structure of the proposed SADAM. It extracts intra-layer feature pairs using multi-branch group convolutions, computes their differences, refines features via spatial-channel attention (SCA), and fuses multi-scale features with residual connections to enhance change detection accuracy.}
\label{fig:SADAM}
\vspace{-4px}
\end{figure}

To adapt the pretrained Hiera-T, we freeze its parameters and insert adapters before each multi-scale block for parameter-efficient fine-tuning~\cite{xiong2024sam2}. Each adapter has a downsampling linear layer, a GeLU activation, an upsampling linear layer, and a final GeLU activation. 
During integration, we discard less relevant SAM2 components like memory attention, prompt encoder, memory encoder, and memory bank.

\textbf{Scale-Aware Differential Attention Module (SADAM).}
The SADAM effectively boosts the network's perception of change objects across diverse scales. As depicted in Fig.~\ref{fig:SADAM}, it utilizes multi-branch group convolutions and spatial-channel attentions to generate high-quality change features. 
Specifically, multiple group convolutions with different receptive fields are employed to extract intra-layer feature pairs $\mathbf{C}^A_{j}$ and $\mathbf{C}^B_{j}$, whose differences are then computed as initial change features $\mathbf{D}_{j}$. The process is expressed as: 
\begin{equation}
	\mathbf{C}^A_{j}, \mathbf{C}^B_{j} = \text{CBR}_{k,g}(\mathbf{C}^A),  \text{CBR}_{k,g}(\mathbf{C}^B), j=1,2,3,4 
\end{equation}
\begin{equation}
	\mathbf{D}_{j} = \text{CBR}_3(|\mathbf{C}^A_{j}- \mathbf{C}^B_{j} |), j=1,2,3,4 
\end{equation}
Here, $\text{CBR}_{k,g}(\cdot)$ denotes a convolution layer with kernel \(k \times k\) and group $g$ (default $g=1$), followed with Batch Normalization and ReLU, where $k=2j-1$, $g=2^j$. The symbol $|\cdot|$ indicates the absolute operation.

% SSCD framework
\begin{figure}[!htb]
\graphicspath{{Fig/}}
\centering
{\includegraphics[width=1.0\linewidth]{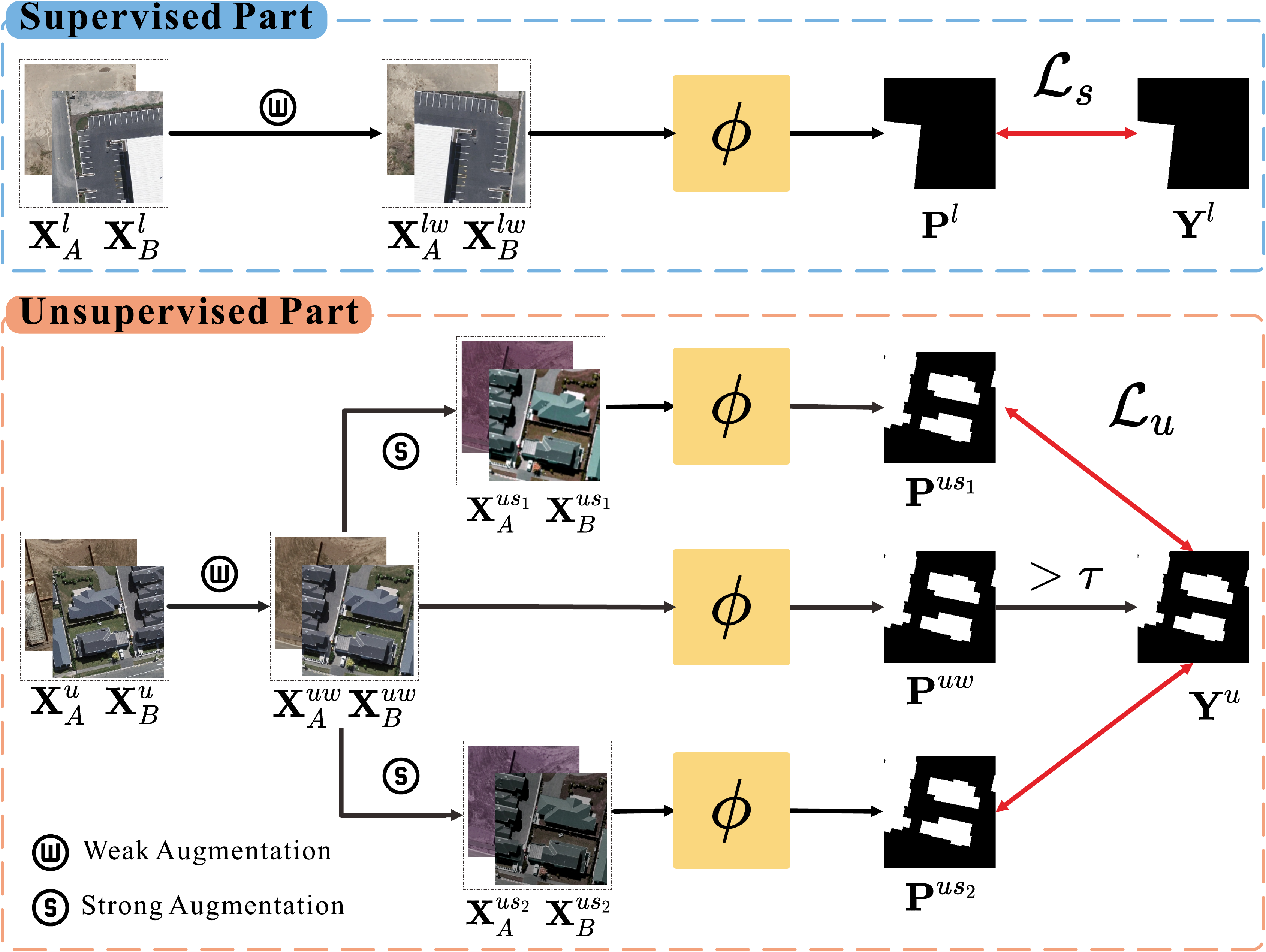}}
\caption{Overall architecture of HSACNet. The framework includes supervised (utilizes labeled data and cross-entropy loss) and unsupervised learning (employs dual-augmentation strong-to-weak consistency for unlabeled data). Here, $\phi$ denotes the proposed change detection network.}
\label{fig:sscdFramework}
\vspace{-6px}
\end{figure}

Next, the spatial-channel attention (SCA)~\cite{song2022fully} is employed on $\mathbf{D}_{j}$ to generate the refined $\mathbf{D}^{SCA}_{j}$, adaptively emphasizing crucial change regions while reducing background noise.
% The details are shown in Appendix~\cref{fig:supplementary_attn}.
It is formulated as:
\begin{equation}
	\mathbf{D}^{SCA}_{j} = \mathbf{D}_{j}+ \gamma \cdot( \sum_{m=1}^{C} A_{m,j} \cdot V_j ), j=1,2,3,4 
\end{equation}
where $\gamma$ is a learnable parameter and $\mathbf{D}_j \in \mathbb{R}^{C \times H \times W}$, with $C$ for channels, $H$ for height, and $W$ for width.   
The features $Q$, $K$, and $V$ are constructed from $\mathbf{D}_j$ with $Q=K=V \in \mathbb{R}^{(H+W)\times C \times S}$, where $S=H=W$. 
The attention weight matrix $A$ is calculated as:
\begin{equation}
	A_{m,n} = \frac{exp(Q_m \cdot K^T_n)}{ {\textstyle \sum_{m=1}^{C}exp(Q_m \cdot K^T_n)} }, m,n \in {1,2,...,C} 
\end{equation}
where $A_{m,n}$ reflects channel-wise relative attention weights.

Subsequently, the features $\mathbf{D}^{SCA}_{j} (j=1,2,3,4)$ are concatenated to form $\mathbf{D}^{cat}$, and a residual change feature $\mathbf{D}^{res}$ is computed. The details are as follows:
\begin{equation}
	\mathbf{D}^{cat} =[\mathbf{D}^{SCA}_{1}, \mathbf{D}^{SCA}_{2}, \mathbf{D}^{SCA}_{3}, \mathbf{D}^{SCA}_{4}]
\end{equation}
\begin{equation}
	\mathbf{D}^{res} = \text{CBR}_3(\text{CBR}_3(|\textbf{C}^A - \textbf{C}^B| ))
\end{equation}
where $[\cdot,\cdot]$ denotes the channel-wise concatenation.
The refined change feature $\textbf{D}^{SADAM}$ is obtained by:
\begin{equation}
	\mathbf{D}^{SADAM} = \text{CBR}_1( \mathbf{D}^{res} + \text{CBR}_1(\mathbf{D}^{cat} ))
\end{equation}
where $\text{CBR}_1(\cdot)$ denotes a \(1 \times 1\) convolution layer with Batch Normalization and ReLU.

\textbf{Decoder.} 
The decoder serves to fuse cross-scale change features and produce change maps. It iteratively combines $\mathbf{D}^{SADAM}_i$ with the upsampled feature, and then utilizes a convolution layer to generate $\mathbf{P}$. 
The formulas are as follows:
\begin{equation}
	\mathbf{F}_{i}= \text{CBR}_3(\text{CBR}_3([\mathbf{D}^{SADAM}_{i}, \text{Up}(\mathbf{F}_{i+1})])), i=3,2,1
\end{equation}
\begin{equation}
	\mathbf{P} = \text{Conv}_1(\mathbf{F}_{1})
\end{equation}
where $\mathbf{F}_4 = \mathbf{D}^{SADAM}_4$, and $\text{Conv}_1$ denotes a \(1 \times 1\) convolution layer. $\text{Up}(\cdot)$ is upsampling operation.

\subsection{Our Proposed HSACNet}
Our HSACNet, illustrated in Fig.~\ref{fig:sscdFramework}, consists of a supervised part and an unsupervised part.

For the supervised part, the labeled image pairs from $\mathcal{D}_l$ are input into the CD network $\phi$. It processes weakly augmented image pairs  $(\mathbf{X}^{lw}_{A},\mathbf{X}^{lw}_{B})$ and generates change maps $\mathbf{P}^l$. We adopt the standard cross-entropy (CE) loss for supervision. The supervised loss $\mathcal{L}_{s}$ is expressed as:
\begin{equation}
	\mathcal{L}_{s} = \mathcal{L}_{CE}(\mathbf{P}^l,\mathbf{Y}^l)
\end{equation}
where $\mathcal{L}_{CE}$ denotes the CE loss.
%which measures the dissimilarity between the predicted change map $\mathbf{P}^l$ and the ground truth $\mathbf{Y}^l$.

For the unsupervised part, we utilize a dual-augmentation strong-to-weak consistency to train $\phi$ with the unlabeled data $\mathcal{D}_u$.
Specifically, two strong augmentations are applied on $(\mathbf{X}^{uw}_{A},\mathbf{X}^{uw}_{B})$, yielding two strong augmented image pairs $(\mathbf{X}^{us_1}_{A},\mathbf{X}^{us_1}_{B})$ and $(\mathbf{X}^{us_2}_{A},\mathbf{X}^{us_2}_{B})$.
They are fed into $\phi$ and generate three change maps $\mathbf{P}^{uw}$, $\mathbf{P}^{us_1}$, and $\mathbf{P}^{us_2}$.
To ensure strong-to-weak consistency, we generate pseudo labels $\mathbf{Y}^u$ from $\mathbf{P}^{uw}$ as follows:
\begin{equation}
\mathbf{Y}^u  = 
	\begin{cases}
		1, & if \ \mathbf{P}^{uw} > \tau \\
		0, & else
	\end{cases}
\end{equation}
where $\tau$ = 0.95 serves as a confidence threshold. The unsupervised loss $\mathcal{L}_u$ is calculated as:
\begin{equation}
	\mathcal{L}_u = \frac{1}{2} ( \mathcal{L}_{CE}(\mathbf{P}^{us_1},\mathbf{Y}^u) + \mathcal{L}_{CE}(\mathbf{P}^{us_2},\mathbf{Y}^u) )
\end{equation}

% main result
\begin{table*}[!ht]
% \small
\centering
\caption{Quantitative comparison on WHU-CD and LEVIR-CD datasets. The highest scores are \textbf{bolded}, and the second \underline{underlined}. }
\vspace{-4px}
\resizebox{0.95\linewidth}{!}{
\label{table:main_result1}
		\begin{tabular}{c|cc|cc|cc|cc|cc|cc|cc|cc}
			\toprule
			% \hline
			\multirow{3}{*}{Method} & \multicolumn{8}{c|}{\textbf{WHU-CD}} & \multicolumn{8}{c}{\textbf{LEVIR-CD}} \\ 
			\multirow{2}{*}{} & \multicolumn{2}{c|}{5\%} & \multicolumn{2}{c|}{10\%} & \multicolumn{2}{c|}{20\%} & \multicolumn{2}{c|}{40\%} & \multicolumn{2}{c|}{5\%} & \multicolumn{2}{c|}{10\%} & \multicolumn{2}{c|}{20\%} & \multicolumn{2}{c}{40\%} \\ 
			& $IoU^c$ & $OA$ & $IoU^c$ & $OA$ & $IoU^c$ & $OA$ & $IoU^c$ & $OA$ & $IoU^c$ & $OA$ & $IoU^c$ & $OA$ & $IoU^c$ & $OA$ & $IoU^c$ & $OA$ \\
			\midrule
			% \hline
			AdvEnt~\cite{vu2019advent} & 57.7 &	97.87 &	60.5 & 97.79 & 69.5 & 98.50 & 76.0 & 98.91 
			& 67.1 & 98.15 & 70.8 & 98.38 & 74.3 & 98.59 & 75.9 & 98.67  \\ 
			s4GAN~\cite{mittal2019semi} & 57.3 & 97.94 & 58.0 & 97.81 & 67.0 & 98.41 & 74.3 & 98.85 
			& 66.6 & 98.16 & 72.2 & 98.48 & 75.1 & 98.63 & 76.2 & 98.68  \\
			SemiCDNet~\cite{peng2020semicdnet} & 56.2 & 97.78 & 60.3 & 98.02 & 69.1 & 98.47 & 70.5 & 98.59 
			& 67.4 & 98.11 & 71.5 & 98.42 & 74.9 & 98.58 & 75.5 & 98.63  \\
			SemiCD~\cite{bandara2022revisiting} & 65.8 & 98.37 & 68.0 & 98.45 & 74.6 & 98.83 & 78.0 & 99.01 
			& 74.2 & \underline{98.59} & 77.1 & 98.74 & 77.9 & 98.79 & 79.0 & 98.84  \\
			RC-CD~\cite{wang2022reliable} & 58.0 & 98.01 & 61.7 & 98.00 & 74.0 & 98.83 & 73.9 & 98.85
			& 74.0 & 98.52 & 76.1 & 98.65 & 77.1 & 98.70 & 77.6 & 98.72  \\
			SemiPTCD~\cite{mao2023semi} & 74.1 & 98.85 & 74.2 & 98.86 & 76.9 & 98.95 & 80.8 & 99.17 
			& 71.2 & 98.39 & 75.9 & 98.65 & 76.6 & 98.65 & 77.2 & 98.74  \\
			UniMatch~\cite{yang2023revisiting} & 78.7 & \underline{99.11} & 79.6 & 99.11 & 81.2 & 99.18 & 83.7 & 99.29 
			& \underline{82.1} & \textbf{99.03} & \underline{82.8} & \textbf{99.07} & \underline{82.9} & 99.07 & 83.0 & \underline{99.08}  \\
			CBFF~\cite{xing2024cbff} & \underline{79.0} & \underline{99.11} & \underline{80.5} & \underline{99.15} & \underline{82.0} & \underline{99.23} & 82.5 & 99.26 
			& \underline{82.1} & \textbf{99.03} & \underline{82.8} & \underline{99.06} & \textbf{83.2} & \textbf{99.09} & \underline{83.3} & \underline{99.08}  \\
			
			\midrule
			% \hline
			Sup-only & 56.4 & 97.90 & 66.1 & 98.38 & 74.4 & 98.82 & \underline{84.3} & \underline{99.34}
			& 72.0 & 98.38 & 77.1 & 98.72 & 81.1 & 98.96 & 82.2 & 99.03 \\
			Ours & \textbf{81.7} & \textbf{99.23} & \textbf{81.3} & \textbf{99.20} & \textbf{84.9} & \textbf{99.35} & \textbf{86.3} & \textbf{99.42}
			& \textbf{82.2} & \textbf{99.03} & \textbf{83.1} & \textbf{99.07} & \textbf{83.2} & \underline{99.08} & \textbf{83.5} & \textbf{99.10} \\
			
			\midrule
			Sup-fully 
			& \multicolumn{8}{c|}{$IoU^c$ = 89.5, $OA$ = 99.27} 
			& \multicolumn{8}{c}{$IoU^c$ = 83.8, $OA$ = 99.11} \\
			
			\bottomrule
	\end{tabular}}
\vspace{-4px}
\end{table*}

\begin{table*}[!ht]
% \small
\centering
\caption{Quantitative comparison on GZ-CD and EGY-BCD datasets. The highest scores are \textbf{bolded}, and the second \underline{underlined}.}
\vspace{-4px}
	\resizebox{0.95\linewidth}{!}{
		\label{table:main_result2}
		\begin{tabular}{c|cc|cc|cc|cc|cc|cc|cc|cc}
			\toprule
			% \hline
			\multirow{3}{*}{Method} & \multicolumn{8}{c|}{\textbf{GZ-CD}} & \multicolumn{8}{c}{\textbf{EGY-BCD}} \\ 
			\multirow{2}{*}{} & \multicolumn{2}{c|}{5\%} & \multicolumn{2}{c|}{10\%} & \multicolumn{2}{c|}{20\%} & \multicolumn{2}{c|}{40\%} & \multicolumn{2}{c|}{5\%} & \multicolumn{2}{c|}{10\%} & \multicolumn{2}{c|}{20\%} & \multicolumn{2}{c}{40\%} \\ 
			& $IoU^c$ & $OA$ & $IoU^c$ & $OA$ & $IoU^c$ & $OA$ & $IoU^c$ & $OA$ & $IoU^c$ & $OA$ & $IoU^c$ & $OA$ & $IoU^c$ & $OA$ & $IoU^c$ & $OA$ \\
			\midrule
			AdvEnt~\cite{vu2019advent} & 56.7 & 95.52 & 57.5 & 95.99 & 70.3 & 97.28 & 70.8 & 97.29 
			& 52.0 & 95.30 & 58.1 & 96.26 & 59.8 & 96.46 & 63.8 & 96.94  \\
			s4GAN~\cite{mittal2019semi} & 59.4 & 96.13 & 61.6 & 96.23 & 68.5 & 97.10 & 69.4 & 97.08 
			& 53.2 & 95.62 & 56.5 & 96.26 & 59.4 & 96.62 & 64.1 & 96.83  \\
			SemiCDNet~\cite{peng2020semicdnet} & 57.9 & 95.38 & 54.9 & 95.52 & 68.9 & 97.16 & 69.7 & 97.20 
			& 52.7 & 95.36 & 56.9 & 96.02 & 59.8 & 96.53 & 63.6 & 96.96  \\
			SemiCD~\cite{bandara2022revisiting} & 59.5 & 96.27 & 58.6 & 96.03 & 67.0 & 97.03 & 71.5 & 97.36 
			& 54.3 & 95.79 & 59.2 & 96.29 & 61.8 & 96.61 & 65.4 & 96.96  \\
			RC-CD~\cite{wang2022reliable} & 62.2 & 96.26 & 63.9 & 96.55 & \underline{74.1} & \underline{97.69} & 74.2 & 97.57 
			& 59.0 & 96.17 & 61.6 & 96.51 & 64.6 & 96.79 & 67.7 & 97.09  \\
			UniMatch~\cite{yang2023revisiting} & \underline{68.7} & \underline{97.06} & \underline{69.5} & \underline{97.41} & 72.8 & 97.71 & 71.1 & 97.48 
			& 62.8 & \underline{96.74} & \underline{65.5} & \underline{97.10} & 63.6 & 96.91 & 67.3 & 97.26 \\
			CBFF~\cite{xing2024cbff} & 67.4 & 97.05 & 69.3 & 97.24 & 72.0 & 97.62 & 76.4 & 97.99 
			& \underline{63.7} & 96.64 & 64.3 & 96.95 & 63.8 & \underline{96.95} & 67.7 & 97.21 \\
			
			\midrule
			Sup-only & 64.8 & 96.51 & 64.2 & 96.75 & 69.8 & 97.23 & \underline{78.2} & \underline{98.08} 
			& 58.2 & 96.05 & 59.6 & 96.26 & \underline{65.0} & \underline{96.95} & \underline{69.1} & \underline{97.35}  \\
			Ours & \textbf{72.5} & \textbf{97.44} & \textbf{73.1} &\textbf{ 97.64} &\textbf{ 80.1} & \textbf{98.27} & \textbf{81.4} & \textbf{98.40}
			& \textbf{68.5} & \textbf{97.25} & \textbf{68.1} & \textbf{97.28} & \textbf{69.6} & \textbf{97.43} & \textbf{70.6} & \textbf{97.50}  \\
			
			\midrule
			Sup-fully 
			& \multicolumn{8}{c|}{$IoU^c$ = 83.3, $OA$ = 98.54} 
			& \multicolumn{8}{c}{$IoU^c$ = 71.7, $OA$ = 97.62} \\
			
			\bottomrule
			
	\end{tabular}}
	\vspace{-6px}
\end{table*}

\textbf{Loss Function.}
The overall loss comprises the supervised loss $\mathcal{L}_s$ and the unsupervised loss $\mathcal{L}_u$. It is formulated as:
\begin{equation}
	\mathcal{L} = \lambda_1 \mathcal{L}_{s} + \lambda_2 \mathcal{L}_{u}
\end{equation}
where $\lambda_1$ = 0.5 and $\lambda_2$ = 0.5.

% %－－－－－－－－－－－－－－－－－－－－－－－－－－－－－－－－－－－－－－－－－－－－－－－－－－－－－－－－－－－－－－

\section{Experiments}

\subsection{Experimental Setups}

\textbf{Datasets.} 
We conduct experiments on four benchmark remote sensing CD datasets, namely WHU-CD~\cite{ji2018fully}, LEVIR-CD~\cite{chen2020spatial}, GZ-CD~\cite{peng2020semicdnet}, and EGY-BCD~\cite{holail2023afde}.
%
%Table~\ref{tab:dataset} details these datasets, including number of image pairs, image size, train/val/test splits, spatial resolution, and change types. 
WHU-CD~\cite{ji2018fully} has two VHR aerial image pairs (2012 and 2016, Christchurch, New Zealand), with \(32507 \times 15354\) pixels and 0.075 m pixel resolution. 
LEVIR-CD contains 637 VHR image pairs (\(1024 \times 1024\) pixels, 0.5 m resolution), showing land use changes (5-14 years, Texas, USA). 
GZ-CD offers 19 high-resolution, season-varying image pairs (2006-2019, Guangzhou suburbs), sized \(1006 \times 1168\) to \(4936 \times 5224\) pixels and 0.55 m resolution. 
EGY-BCD comprises 6091 VHR image pairs (2017-2022, four Egyptian regions), \(256 \times 256\) pixels and 0.25 m resolution.
All images are cropped into \(256\times 256\) non-overlapping patches and split into training, validation, and test sets. 
The training set is further divided into labeled and unlabeled data at ratios of \([5\%,95\%]\), \([10\%,90\%]\), \([20\%,80\%]\), \([40\%,60\%]\).

\textbf{Baselines.} 
%To validate the effectiveness and superiority of our method, 
We compare our method with eight existing state-of-the-art methods: AdvEnt~\cite{vu2019advent}, s4GAN~\cite{mittal2019semi}, SemiCDNet~\cite{peng2020semicdnet}, SemiCD~\cite{bandara2022revisiting}, RC-CD~\cite{wang2022reliable}, SemiPTCD~\cite{mao2023semi}, UniMatch~\cite{yang2023revisiting}, and CBFF~\cite{xing2024cbff}.
Among them, the first three are GAN-based SSCD methods, RC-CD is a pseudo-label-based method, and the remaining four are consistency regularization-based methods.
Sup-only denotes our method trained only on labeled data, while Sup-fully denotes our method trained with all training labels.
All methods are implemented with PyTorch and trained on the same training sets.

\textbf{Metrics.} 
We adopt intersection over union ($IoU^c$) and overall accuracy ($OA$) to evaluate performance. $IoU^c$ quantifies precision by measuring the overlap between predicted and actual changes, while $OA$ offers a general accuracy measure. 
Their formulas are $IoU^c = \frac{TP}{TP + FP + FN}$, $OA = \frac{TP + TN}{TP + FP + FN + TN}$,
with $TP$, $TN$, $FP$ and $FN$ signifying true positive, true negative, false positive, and false negative.

\begin{figure*}[!ht]
	\graphicspath{{Fig/}}
	\centering
	{\includegraphics[width=0.80\linewidth]{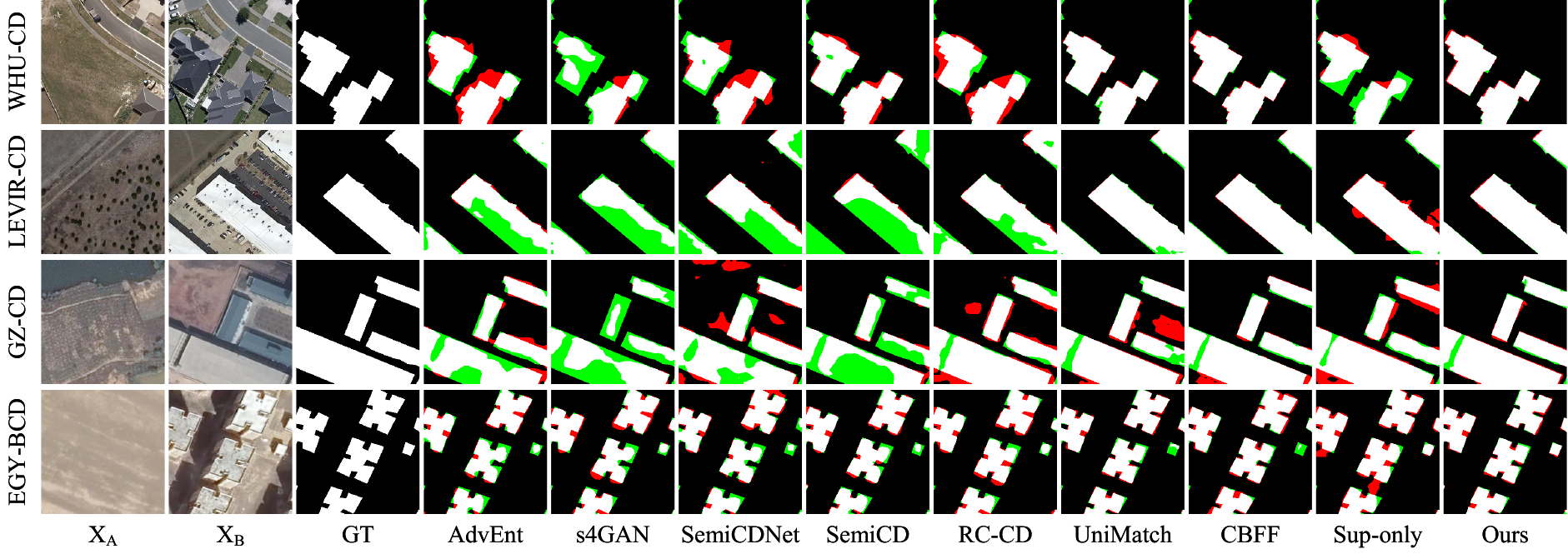}}
	\vspace{-2px}
	\caption{Detection results of different methods on four CD datasets at \(5\%\) labeled ratio. Here, white for $TP$, black for $TN$, red for $FP$, and green for $FN$. }
	\vspace{-8px}
	\label{fig:result_visual}
\end{figure*}

\textbf{Implementation Details.} 
The experiments are conducted with PyTorch on an NVIDIA RTX2080Ti GPU. We adopt the AdamW optimizer with an initial learning rate of 3e-4, momentum of 0.9, and weight decay of 1e-6. The CosineAnnealingLR scheduler is utilized to improve training stability and convergence, annealing the learning rate to 1e-6 within 80 epochs. The training process spans 80 epochs with a batch size of 4.
To enrich training data and boost model capabilities, we employ both weak and strong augmentation techniques. Weak augmentations involve random resizing and horizontal flipping, while strong augmentations include random color jittering, Gaussian blur, and CutMix.

\subsection{Main Results}

\textbf{Comparison with the State-of-the-Art.} 
Table~\ref{table:main_result1} and \ref{table:main_result2} present quantitative comparisons of different methods on WHU-CD, LEVIR-CD, GZ-CD, and EGY-BCD datasets.
From these tables, we can observe:
\ding{182} Our method demonstrates the best performance across all datasets and labeled proportions. 
On WHU-CD, compared to CBFF, our method gains \(2.7\%\), \(0.8\%\), \(2.9\%\), and \(3.8\%\) in $IoU^c$ with \(5\%\), \(10\%\), \(20\%\), and \(40\%\) labeled data.
On GZ-CD, the $IoU^c$ improvements of our method over UniMatch are \(3.8\%\), \(3.6\%\), \(7.3\%\), and \(8.5\%\).
On EGY-BCD, the corresponding $IoU^c$ improvements are \(4.8\%\), \(3.8\%\), \(5.8\%\), and \(2.9\%\).
\ding{183} Our sup-only variant has a remarkable performance. As labeled ratio increases, it exceeds most semi-supervised methods. On \(20\%\) and \(40\%\) partitions of EGY-BCD, it reaches $IoU^c$ values of 65.0 and 69.1, surpassing competitors.
\ding{184} Our sup-fully variant shows a outstanding performance, serving as an upper bound of our CD network. 
These quantitative results highlight our method's robustness and adaptability, achieving excellent performance across various datasets and labeled proportions.

\textbf{Result Visualization.}
Fig.~\ref{fig:result_visual} shows some typical examples of different methods on four CD datasets under \(5\%\) labeled partition. It is evident that our method exhibits higher accuracy and richer details. Both quantitative and qualitative results verify the superiority of our method.
% results support our method's superiority.

\textbf{Computational Complexity Analysis.}
Table~\ref{tab:comparison} compares parameters and computational costs among different methods under \(256 \times 256\) image size setting.
Our method outperforms others in computational efficiency, with only 17.55 GFLOPs and 28.9 M parameters. Compared with UniMatch and CBFF, our method significantly reduces computational costs and model size, confirming its effectiveness and efficiency.

\vspace{-4px}
\begin{table}[!ht]
	\centering
	\caption{Comparison of parameters and computational costs.}
	\vspace{-4px}
	\label{tab:comparison}
	\resizebox{0.26\textwidth}{!}{
		\large
		\begin{tabular}{c | c c}
			\toprule
			Method & GFLOPs & Params (M) \\
			\midrule
			AdvEnt & 74.06 & 49.6   \\
			s4GAN & 76.72 & 49.6   \\
			SemiCDNet & 75.69 & 52.4  \\
			SemiCD & 75.37 & 50.7   \\
			UniMatch & 38.30 & 40.5  \\
			CBFF & 27.24 & 46.4   \\
			\midrule
			% GTPC & 65.87 & 57.3   \\
			Ours & \textbf{17.55} & \textbf{28.9}  \\
			\bottomrule
	\end{tabular}}
\vspace{-8px}
\end{table}

\subsection{Ablation Studies}

\textbf{Component Effectiveness Analysis.}
To explore the contributions of each component, we conduct a set of ablation studies. 
Table~\ref{tab:module_comparison} shows performance and cost comparisons of different variants on \(5\%\) labeled WHU-CD.
The ``w/o SAM2'' variant, replacing SAM2 with Resnet50, has an $IoU^c$ of 79.3 and has 27.6 M parameters. The "w/o SADAM" setting, using a same-parameter convolution layer instead of SADAM, attains an $IoU^c$ of 77.1 and has 28.9 M parameters. In contrast, our method outperforms both, reaching an $IoU^c$ of 81.7 and has 28.9 M parameters.
This comparison clearly demonstrates the effectiveness of each component.

\begin{table}[!ht]
\vspace{-4px}
\centering
\caption{Module variants comparison on \(5\%\) labeled WHU-CD.}
\vspace{-4px}
\label{tab:module_comparison}
\resizebox{0.37\textwidth}{!}{
%\large
\begin{tabular}{c | c c c c}
	\toprule
	Method & $IoU^c$ & $OA$ & GFLOPs & Params (M) \\
	\midrule
	Base & 77.9 & 99.03 & 16.32 & 26.3 \\
	w/o SAM2 & 79.3 & \textbf{99.23} & 17.05 & 27.6 \\
	w/o SADAM & 77.1 & 99.04 & 16.97 & 28.9   \\
	\midrule
	Ours & \textbf{81.7} & \textbf{99.23} & 17.55 & 28.9  \\
	\bottomrule
\end{tabular}}
\vspace{-4px}
\end{table}

\begin{figure}[!ht]
\vspace{-8px}
\graphicspath{{Fig/}}
\centering
{\includegraphics[width=0.95\linewidth]{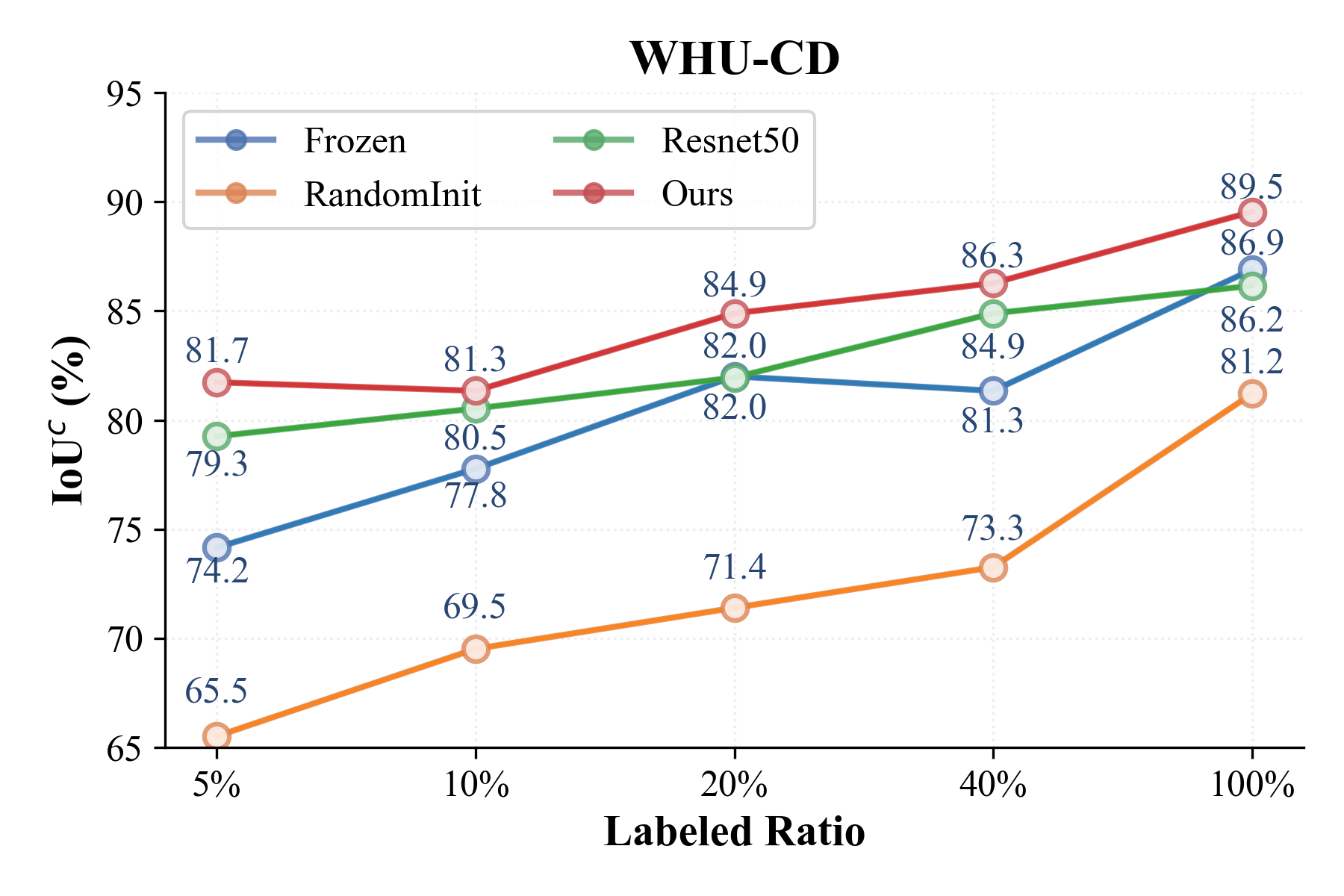}}
\vspace{-8px}
\caption{Performance comparison and trend of multiple approaches with varying labeled ratios on WHU-CD dataset. }
\vspace{-4px}
\label{fig:SAM2Adapter_ablation2}
\end{figure}

\textbf{The Effectiveness of SAM2 Adapter Fine-tuning.}
Fig.~\ref{fig:SAM2Adapter_ablation2} depicts the performance comparisons of multiple variants on WHU-CD (labeled ratios on x-axis, $IoU^c$ on y-axis). 
The ``Frozen" variant of keeping pretrained weights without adapter fine-tuning, the ``RandomInit" variant starting with random weights and then fine-tuning, and the ``Resnet50" variant replacing SAM2 with Resnet50, all perform worse than our method.
Across labeled data ratios, our method exhibits higher $IoU^c$ than other three, showing a distinct advantage. The stability of its line as the labeled ratio increases reflects its robustness and adaptability.
The performance validates the effectiveness of SAM2 adapter fine-tuning in SSCD and its adaptability to varying labeled data amounts.

\begin{figure}[!ht]
	\vspace{-12px}
	\graphicspath{{Fig/}}
	\centering
	{\includegraphics[width=0.90\linewidth]{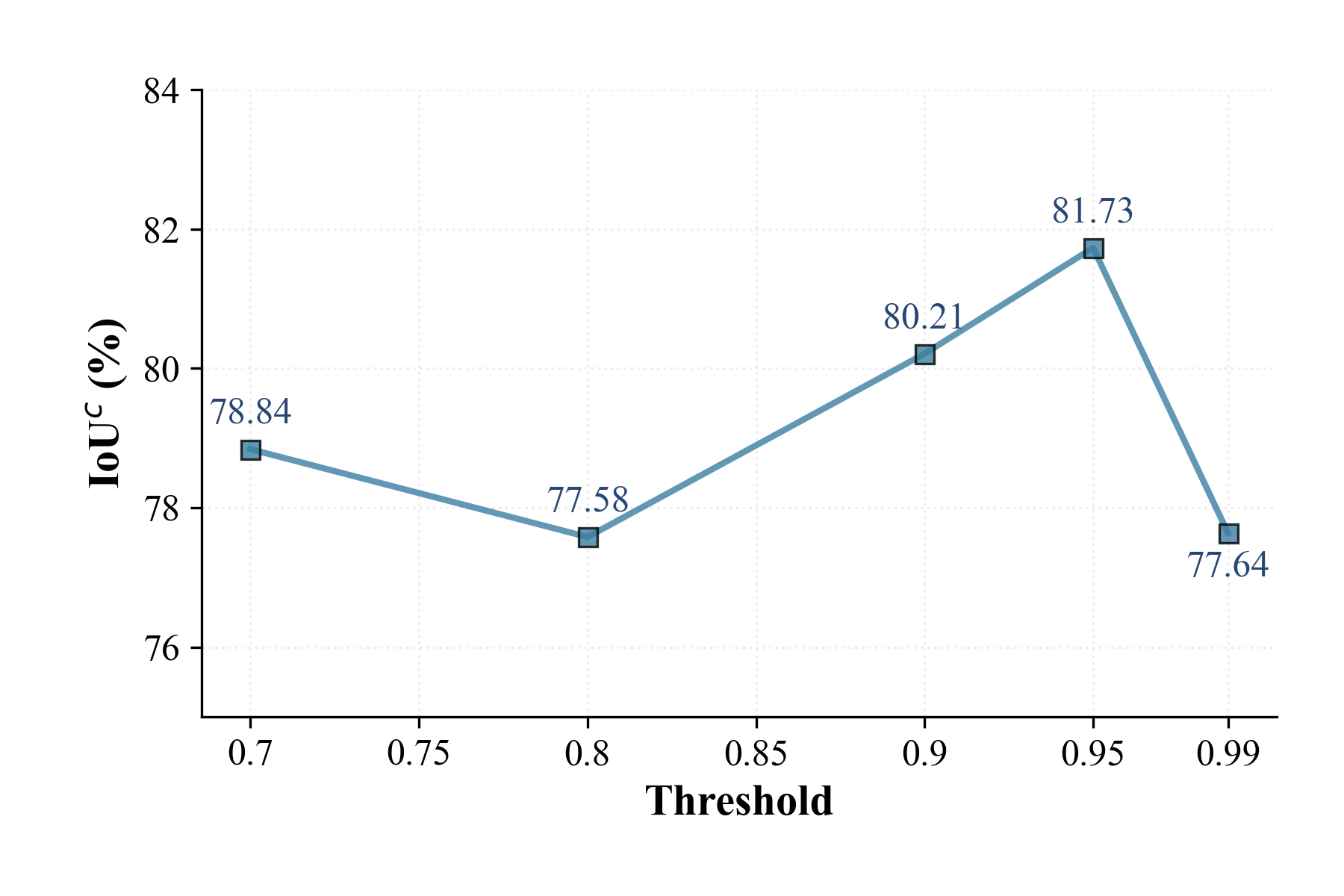}}
	\vspace{-12px}
	\caption{$IoU^c$ of \(5\%\) labeled WHU-CD at different thresholds. }
	\vspace{-8px}
	\label{fig:Threshold}
\end{figure}

\textbf{Threshold Sensitivity Analysis.}
We add sensitivity experiments regarding threshold $\tau$ to access the performance changes of our method with varying thresholds. Fig.~\ref{fig:Threshold} clearly shows that the performance varies in accordance with different threshold values, attaining its peak when $\tau$ is 0.95. 

%

% %－－－－－－－－－－－－－－－－－－－－－－－－－－－－－－－－－－－－－－－－－－－－－－－－－－－－－－－－－－－－－－

\section{Conclusion}

In this paper, we propose HSACNet, a Hierarchical Scale-Aware Consistency regularized Network for semi-supervised change detection. 
By integrating SAM2's Hiera backbone and applying adapters for parameter-efficient fine-tuning, it effectively extracts inter-layer multi-scale features. 
The SADAM harnesses multi-branch group convolutions and spatial-channel attention to capture intra-layer multi-scale features and suppress noise interference. 
Meanwhile, we utilize dual-augmentation consistency regularization to mine rich information from unlabeled data.
We have evaluated HSACNet on four benchmark datasets, with the results validating its superiority and effectiveness.

\bibliographystyle{ieeetr}
\bibliography{icme2025}

\begin{thebibliography}{10}

\bibitem{bai2023deep}
T.~Bai, L.~Wang, D.~Yin, K.~Sun, Y.~Chen, W.~Li, and D.~Li, ``Deep learning for change detection in remote sensing: a review,'' {\em Geo-spatial Information Science}, vol.~26, no.~3, pp.~262--288, 2023.

\bibitem{Holail2024nsr}
S.~Holail, T.~Saleh, X.~Xiao, J.~Xiao, G.-S. Xia, Z.~Shao, J.~Gong, and D.~Li, ``Time-series satellite remote sensing reveals gradually increasing war damage in the gaza strip,'' {\em National Science Review}, vol.~11, p.~nwae304, 08 2024.

\bibitem{cheng2023change}
G.~Cheng, Y.~Huang, X.~Li, S.~Lyu, Z.~Xu, H.~Zhao, Q.~Zhao, and X.~Shiming, ``Change detection methods for remote sensing in the last decade: A comprehensive review,'' {\em Remote Sensing}, vol.~16, no.~13, p.~2355, 2024.

\bibitem{ravi2024sam}
N.~Ravi, V.~Gabeur, Y.-T. Hu, R.~Hu, C.~Ryali, T.~Ma, H.~Khedr, R.~R{\"a}dle, C.~Rolland, L.~Gustafson, {\em et~al.}, ``{SAM} 2: Segment anything in images and videos,'' in {\em The Thirteenth International Conference on Learning Representations}, 2025.

\bibitem{peng2020semicdnet}
D.~Peng, L.~Bruzzone, Y.~Zhang, H.~Guan, H.~Ding, and X.~Huang, ``Semicdnet: A semisupervised convolutional neural network for change detection in high resolution remote-sensing images,'' {\em IEEE Transactions on Geoscience and Remote Sensing}, vol.~59, no.~7, pp.~5891--5906, 2020.

\bibitem{wang2022reliable}
J.-X. Wang, T.~Li, S.-B. Chen, J.~Tang, B.~Luo, and R.~C. Wilson, ``Reliable contrastive learning for semi-supervised change detection in remote sensing images,'' {\em IEEE Transactions on Geoscience and Remote Sensing}, vol.~60, pp.~1--13, 2022.

\bibitem{yang2023revisiting}
L.~Yang, L.~Qi, L.~Feng, W.~Zhang, and Y.~Shi, ``Revisiting weak-to-strong consistency in semi-supervised semantic segmentation,'' in {\em 2023 IEEE/CVF Conference on Computer Vision and Pattern Recognition (CVPR)}, pp.~7236--7246, 2023.

\bibitem{kirillov2023segment}
A.~Kirillov, E.~Mintun, N.~Ravi, H.~Mao, C.~Rolland, L.~Gustafson, T.~Xiao, S.~Whitehead, A.~C. Berg, W.-Y. Lo, {\em et~al.}, ``Segment anything,'' in {\em Proceedings of the IEEE/CVF International Conference on Computer Vision}, pp.~4015--4026, 2023.

\bibitem{ryali2023hiera}
C.~Ryali, Y.-T. Hu, D.~Bolya, C.~Wei, H.~Fan, P.-Y. Huang, V.~Aggarwal, A.~Chowdhury, O.~Poursaeed, J.~Hoffman, {\em et~al.}, ``Hiera: A hierarchical vision transformer without the bells-and-whistles,'' in {\em International Conference on Machine Learning}, pp.~29441--29454, PMLR, 2023.

\bibitem{xiong2024sam2}
X.~Xiong, Z.~Wu, S.~Tan, W.~Li, F.~Tang, Y.~Chen, S.~Li, J.~Ma, and G.~Li, ``Sam2-unet: Segment anything 2 makes strong encoder for natural and medical image segmentation,'' {\em arXiv preprint arXiv:2408.08870}, 2024.

\bibitem{song2022fully}
Q.~Song, J.~Li, C.~Li, H.~Guo, and R.~Huang, ``Fully attentional network for semantic segmentation,'' in {\em Proceedings of the AAAI Conference on Artificial Intelligence}, vol.~36, pp.~2280--2288, 2022.

\bibitem{vu2019advent}
T.-H. Vu, H.~Jain, M.~Bucher, M.~Cord, and P.~P{\'e}rez, ``Advent: Adversarial entropy minimization for domain adaptation in semantic segmentation,'' in {\em 2019 IEEE/CVF Conference on Computer Vision and Pattern Recognition (CVPR)}, pp.~2517--2526, 2019.

\bibitem{mittal2019semi}
S.~Mittal, M.~Tatarchenko, and T.~Brox, ``Semi-supervised semantic segmentation with high-and low-level consistency,'' {\em IEEE Transactions on Pattern Analysis and Machine Intelligence}, vol.~43, no.~4, pp.~1369--1379, 2019.

\bibitem{bandara2022revisiting}
W.~G.~C. Bandara and V.~M. Patel, ``Revisiting consistency regularization for semi-supervised change detection in remote sensing images,'' {\em arXiv preprint arXiv:2204.08454}, 2022.

\bibitem{mao2023semi}
Z.~Mao, X.~Tong, and Z.~Luo, ``Semi-supervised remote sensing image change detection using mean teacher model for constructing pseudo-labels,'' in {\em ICASSP 2023-2023 IEEE International Conference on Acoustics, Speech and Signal Processing (ICASSP)}, pp.~1--5, 2023.

\bibitem{xing2024cbff}
Y.~Xing, Q.~Xu, J.~Zeng, R.~Huang, S.~Gao, W.~Xu, Y.~Zhang, and W.~Fan, ``Cross branch feature fusion decoder for consistency regularization-based semi-supervised change detection,'' in {\em ICASSP 2024 - 2024 IEEE International Conference on Acoustics, Speech and Signal Processing (ICASSP)}, pp.~9341--9345, 2024.

\bibitem{ji2018fully}
S.~Ji, S.~Wei, and M.~Lu, ``Fully convolutional networks for multisource building extraction from an open aerial and satellite imagery data set,'' {\em IEEE Transactions on Geoscience and Remote Sensing}, vol.~57, no.~1, pp.~574--586, 2018.

\bibitem{chen2020spatial}
H.~Chen and Z.~Shi, ``A spatial-temporal attention-based method and a new dataset for remote sensing image change detection,'' {\em Remote Sensing}, vol.~12, no.~10, p.~1662, 2020.

\bibitem{holail2023afde}
S.~Holail, T.~Saleh, X.~Xiao, and D.~Li, ``Afde-net: Building change detection using attention-based feature differential enhancement for satellite imagery,'' {\em IEEE Geoscience and Remote Sensing Letters}, vol.~20, pp.~1--5, 2023.

\end{thebibliography}

% 补充材料
% \clearpage
% \appendix
% \renewcommand\thetable{\Alph{section}\arabic{table}}  
% \setcounter{table}{0}
% \renewcommand\thefigure{\Alph{section}\arabic{figure}}  
% \setcounter{figure}{0}

%\renewcommand\thetable{\Alph{section}\arabic{table}}  
%\setcounter{table}{0}
%\renewcommand\thefigure{\Alph{section}\arabic{figure}}  
%\setcounter{figure}{0}

% % 
\section{Appendix}

\subsection{Dataset Details}
Appendix~\cref{tab:supplementary_dataset} details the four used CD datasets, including the number of image pairs, image size, train/val/test splits, spatial resolution, and change types. 

%Their descriptions are as follows:
%\begin{itemize}
%	\item WHU-CD comprises two Very High-Resolution (VHR) aerial image pairs acquired in 2012 and 2016 over Christchurch, New Zealand. Each image boasts a resolution of \(32507 \times 15354\) pixels and a pixel resolution of 0.075 m. 
% 	\item LEVIR-CD consists of 637 VHR image pairs with a resolution of \(1024 \times 1024\) pixels and a pixel resolution of 0.5 m, capturing substantial land use alterations over periods ranging from 5 to 14 years across various regions in Texas, USA.
% 	\item GZ-CD presents 19 high-resolution, season-varying image pairs from Guangzhou's suburbs, captured between 2006 and 2019, with resolutions from \(1006 \times 1168\) to \(4936 \times 5224\) pixels and a pixel resolution of 0.55 m.
% 	\item EGY-BCD includes 6091 VHR image pairs from four Egyptian regions, captured between 2017 and 2022, each with a uniform size of \(256 \times 256\) pixels and a pixel resolution of 0.25 m.
%\end{itemize}

%
\begin{table*}[!hbp]
	\centering
	\caption{Dataset details.}
	\label{tab:supplementary_dataset}
	\resizebox{0.85\textwidth}{!}{%
		\begin{tabular}{c|cccc}
			\toprule
			Category & WHU-CD & LEVIR-CD & GZ-CD & EGY-BCD \\
			\midrule
			Image Pairs & 1 & 637 & 19 & 6091 \\
			Image Size & $15354 \times 32507 \times 3$ & $1024 \times 1024 \times 3$ & \makecell{$1006 \times 1168 \times 3$, \\ $4936 \times 5224 \times 3$} & $256 \times 256 \times 3$  \\
			Train/Validation/Test Split & 5974/743/744 & 7120/1024/2048 & 2882/360/361 & 4264/1218/609 \\
			Spatial Resolution & 0.075 m/pixel & 0.5 m/pixel & 0.55 m/pixel & 0.25 m/pixel  \\
			Change Types & Buildings & \makecell{Buildings, villa residences, \\ apartments, small garages, and etc.} & \makecell{Buildings, roads, forests, \\ farmlands, and etc.} & Buildings  \\
			\bottomrule
	\end{tabular}}
\end{table*}

\subsection{Spatial-Channel Attention (SCA)}
\begin{figure}[!htbp]
	\graphicspath{{Fig/}}
	\centering
	{\includegraphics[width=0.95\linewidth]{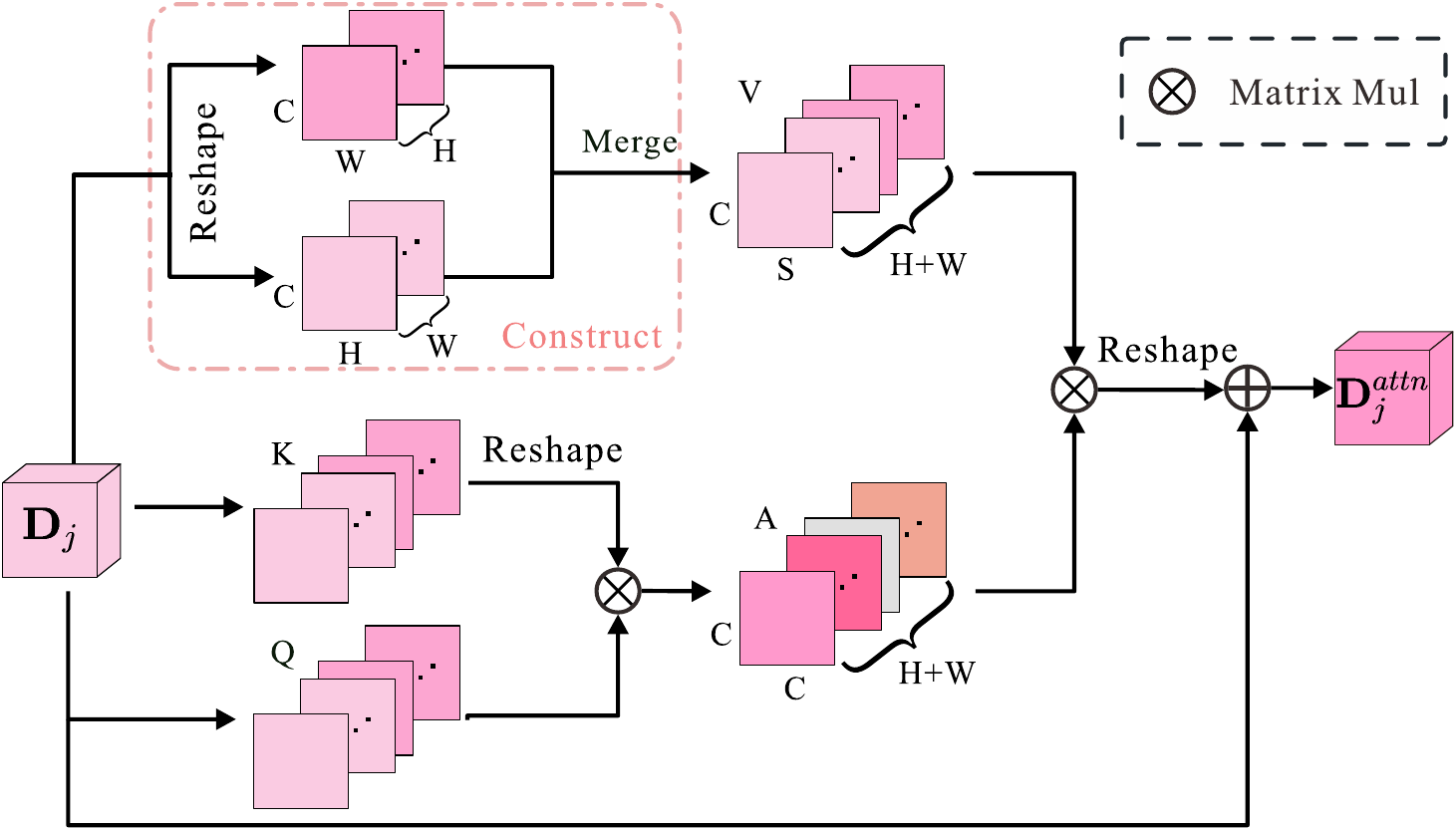}}
	\caption{Structure of the spatial-channel attention (SCA). }
	\label{fig:supplementary_attn}
\end{figure}

Fig.~\ref{fig:supplementary_attn} shows the structure of the spatial-channel attention.
The construction process of \(Q\), \(K\), and \(V\) in the attention mechanism is as follows. The input feature \(\mathbf{D}_j\) is segmented along the height ($H$) and width ($W$) dimensions, yielding two groups of slices: \(H\) slices of size \(C\times W\) and \(W\) slices of size \(C\times H\). Since \(H = W\), these slices are stacked and fused to obtain a new feature $Q=K=V \in \mathbb{R}^{(H+W)\times C \times S}$.
This construction process can enrich the feature representation and assist the model in capturing important information in both spatial and channel dimensions simultaneously.

\subsection{Baselines}
We have selected eight existing state-of-the-art methods, their descriptions are as follows:
\begin{itemize}
	\item AdvEnt enhances domain adaptation by minimizing target domain prediction entropy with adversarial networks, promoting confident target predictions.
	\item s4GAN utilizes unlabeled data through dual-branch architecture and GAN-generated pseudo-labels.
	\item SemiCDNet adopts the UNet++ architecture with attention modules to generate change maps, and incorporates entropy adversarial loss and segmentation adversarial loss to leverage unlabeled data.
	\item SemiCD applies different perturbations to change features and minimizes the difference between unperturbed and perturbed change prediction maps using MSE loss.
	\item RC-CD involves selecting reliable pseudo-labels, retraining the network, and implementing a contrastive loss to improve the feature extraction ability.
	\item SemiPTCD employs a mean-teacher model for strong-to-weak consistency exploration and integrates T-VAT for iterative model refinement and enhanced generalization.
	\item UniMatch exploits unlabeled data by simply incorporating multi-level consistency regularization.
	\item CBFF proposes a cross-branch feature fusion decoder that combines convolution and Transformer, and utilizes unlabeled data through strong-to-weak consistency.
\end{itemize}

\end{document}